\documentclass[12pt,draftcls,onecolumn]{IEEEtran}
\usepackage{amssymb}
\usepackage{amsbsy}
\usepackage{amsmath}
\usepackage{amsthm}
\usepackage{amsfonts}
\usepackage{setspace}
\usepackage{epsfig}
\usepackage{cite}
\usepackage{subfigure}
\usepackage{lscape}
\usepackage{multirow}
\usepackage{placeins}
\usepackage{float}

\numberwithin{equation}{section}

\title{On the Subspace of Image Gradient Orientations}
\author{Georgios Tzimiropoulos and Stefanos Zafeiriou\\
\emph{Imperial College London}}

\begin{document}
%
\maketitle

\begin{abstract}

We introduce the notion of Principal Component Analysis (PCA) of
image gradient orientations. As image data is typically noisy, but
noise is substantially different from Gaussian, traditional PCA of
pixel intensities very often fails to estimate reliably the
low-dimensional subspace of a given data population. We show that
replacing intensities with gradient orientations and the $\ell_2$
norm with a cosine-based distance measure offers, to some extend, a
remedy to this problem. Our scheme requires the eigen-decomposition
of a covariance matrix and is as computationally efficient as
standard $\ell_2$ PCA. We demonstrate some of its favorable
properties on robust subspace estimation.

\end{abstract}

\newpage
\begin{center}
NOTATION
\end{center}
\ \\
\begin{tabular}{ll}
$\mathcal{S}$, $\{.\}$ & \text{set} \\
$\Re$ & set of reals\\
$\mathcal{C}$ & set of complex numbers\\
x & \text{scalar} \\
$\mathbf{x}$ & \text{column vector} \\
$\mathbf{X}$ & \text{matrix} \\
 $\mathbf{I}_{m\times m}$ & $m \times m$ identity matrix\\
$\mathbf{a}(k)$ & $k$-th  element of vector $\mathbf{a}$ \\
$N(\mathcal{S})$ & cardinality set $\mathcal{S}$ \\
$||.||$ &  $\ell_2$ norm\\
$||.||_F$ &  Frobenius norm \\
$\mathbf{Z}^H$ & conjugate transpose of $\mathbf{Z}$ \\
$U[a,b]$ & uniform distribution in $[a,b]$\\
$\mathbb{E}[.]$ & mean value operator \\
$x \sim U[a,b]$ & $x$ follows $U[a,b]$

\end{tabular}

\section{Introduction}

Provision  for mechanisms capable of handling gross
errors caused by possible arbitrarily large model deviations is a
typical prerequisite in computer vision. Such deviations are not
unusual in real-world applications where data contain artifacts due
to occlusions, illumination changes, shadows, reflections or the
appearance of new parts/objects. In most cases, such phenomena
cannot be described by a mathematically well-defined generative
model and are usually referred as outliers in learning and parameter
estimation.

In this paper, we propose a new avenue for Principal Component
Analysis (PCA), perhaps the most classical tool for dimensionality
reduction and feature extraction in pattern recognition. Standard
PCA estimates the $k-$rank linear subspace of the given data
population, which is optimal in a least-squares sense. Unfortunately
$\ell_2$ norm enjoys optimality properties only when noise is i.i.d.
Gaussian; for data corrupted by outliers, the estimated subspace can
be arbitrarily biased.

Robust formulations to PCA, such as robust covariance matrix
estimators \cite{Cam1980,CroHae2000}, are computationally
prohibitive for high dimensional data such as images. Robust
approaches, well-suited for computer vision applications, include
$\ell_1$ \cite{KeKan2005,Kwa2008}, robust energy function
\cite{TorBla2003} and weighted combination of nuclear norm and
$\ell_1$ minimization \cite{ChaSanParWil2009,CanLiMaWri2009}.
$\ell_1$-based approaches can be computational efficient, however
the gain in robustness is not always significant. The M-Estimation
framework of \cite{TorBla2003} is robust but suitable only for
relatively low dimensional data or off-line processing. Under weak
assumptions \cite{CanLiMaWri2009}, the convex optimization
formulation of \cite{ChaSanParWil2009,CanLiMaWri2009} perfectly
recovers the low dimensional subspace of a data population corrupted
by sparse arbitrarily large errors; nevertheless efficient
reformulations of standard PCA can be orders of magnitude faster.

In this paper we look at robust PCA from a completely different
perspective. Our scheme \textit{does not} operate on pixel
intensities. In particular, we replace pixel intensities with
gradient orientations. We define a notion of pixel-wise image
dissimilarity by looking at the distribution of gradient orientation
differences; intuitively this must be approximately uniform in
$[0,2\pi)$. We then assume that local orientation mismatches caused
by outliers can be also well-described by a uniform distribution
which, under some mild assumptions, is canceled out when we apply
the cosine kernel. This last observation has been noticed in
recently proposed schemes for image registration
\cite{TziArgZafSta2010}. Following this line of research, we show
that a cosine-based distance measure has a functional form which
enables us to define an explicit mapping from the space of gradient
orientations into a high-dimensional complex sphere where essentially
linear complex PCA is performed. The mapping is one-to-one and
therefore PCA-based reconstruction in the original input space is
direct and requires no further optimization. Similarly to standard
PCA, the basic computational module of our scheme requires the
eigen-decomposition of a covariance matrix, while high dimensional
data can be efficiently analyzed following the strategy suggested in
Turk and Pentland's Eigenfaces \cite{TurPen1991}.

\section{$\ell_2$-based PCA of pixel intensities}\label{S:PCA}

Let us denote by $\mathbf{x}_i \in \Re^{p}$ the $p-$dimensional
vector obtained by writing image $\mathbf{I}_i \in \Re^{m_1 \times
m_2}$ in lexicographic ordering. We assume that we are given a
population of $n$ samples $\mathbf{X}=[\mathbf{x}_1|\cdots|
\mathbf{x}_n]\in \Re^{p\times n}$. Without loss of generality, we
assume zero-mean data. PCA finds a set of $k<n$ orthonormal bases
$\mathbf{B}_k=[\mathbf{b}_1|\cdots| \mathbf{b}_k]\in \Re^{p\times
k}$ by minimizing the error function
\begin{equation}
\epsilon(\mathbf{B}_k) =
||\mathbf{X}-\mathbf{B}_k\mathbf{B}_k^T\mathbf{X}||^2_F\label{L_2
PCA}.
\end{equation}
The solution is given by the eigenvectors corresponding to the $k$
largest eigenvalues obtained from the eigen-decomposition of the
covariance matrix $\mathbf{X}\mathbf{X}^T$. Finally, the
reconstruction of $\mathbf{X}$ from the subspace spanned by the
columns of $\mathbf{B}_k$ is given by
$\mathbf{\tilde{X}}=\mathbf{B}_k\mathbf{C}_k$, where
$\mathbf{C}_k=\mathbf{B}_k^T\mathbf{X}$ is the matrix which gathers
the set of projection coefficients.

For high dimensional data and Small Sample Size (SSS) problems (i.e.
$n \ll p$), an efficient implementation of PCA in $O(n^3)$ (instead
of $O(p^3)$) was proposed in \cite{TurPen1991}. Rather than
computing the eigen-analysis of $\mathbf{X}\mathbf{X}^T$, we compute
the eigen-analysis of
$\mathbf{X}^T\mathbf{X}$ and make use of the following theorem\\
\textbf{Theorem I}\\
Define matrices $\mathbf{A}$ and $\mathbf{B}$ such that $\mathbf{A}
= \mathbf{\Gamma}\mathbf{\Gamma}^H$ and $\mathbf{B} =
\mathbf{\Gamma}^H\mathbf{\Gamma}$ with $\mathbf{\Gamma}\in
\mathcal{C}^{m \times r}$. Let $\mathbf{U}_A$ and $\mathbf{U}_B$ be
the eigenvectors corresponding to the non-zero eigenvalues
$\mathbf{\Lambda}_A$ and $\mathbf{\Lambda}_B$ of $\mathbf{A}$ and
$\mathbf{B}$, respectively. Then, $\mathbf{\Lambda}_A =
\mathbf{\Lambda}_B$ and $\mathbf{U}_A = \mathbf{\Gamma}\mathbf{U}_B
\mathbf{\Lambda}_A^{-\frac{1}{2}}$.


\section{Random number generation from gradient orientation
differences}\label{S:RNG}

\begin{figure}[ht!]
\centering
\includegraphics[width=0.75\columnwidth]{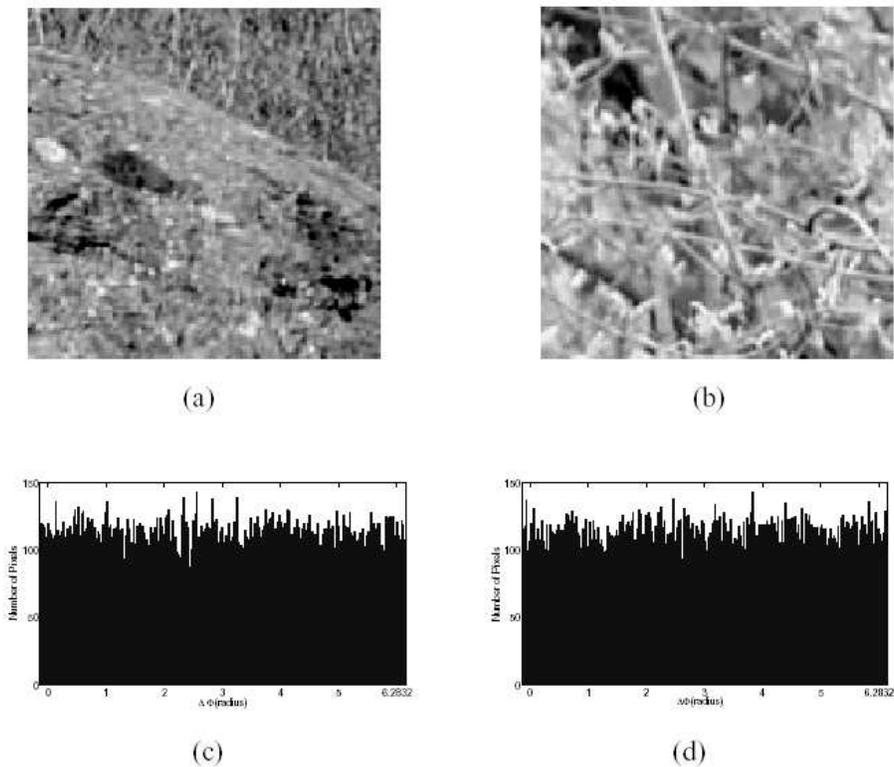}
\caption{(a)-(b) An image pair used in our experiment, (c)
Image-based random number generator: histogram of 40,000 gradient
orientation differences and (d) Histogram of 40,000 samples drawn
from Matlab's random number generator.}\label{Fig1}
\end{figure}

We formalize an observation for the distribution of gradient
orientation differences which does not appear to be well-known in
the scientific community \footnote{This observation has been
somewhat noticed in \cite{FitKadChrKit2002} with no further comments
on its implications.}. Consider a set of images $\{\mathbf{J}_i\}$.
At each pixel location, we estimate the image gradients and the
corresponding gradient orientation \footnote{More specifically, we
compute $\Phi_i = \arctan{G_{i,y}/G_{i,x}}$, where $G_{i,x}=h_x\star
I_i$, $G_{i,y}=h_y\star I_i$ and $h_x,h_y$ are filters used to
approximate the ideal differentiation operator along the image
horizontal and vertical direction respectively. Possible choices for
$h_x,h_y$ include central difference estimators of various orders
and discrete approximations to the first derivative of the
Gaussian.}. We denote by $\{\mathbf{\Phi}_i\}\textrm{,
}\mathbf{\Phi}_i \in [0,2\pi)^{m_1\times m_2}$ the set of
orientation images and compute the orientation difference image
\begin{equation}
\Delta\mathbf{\Phi}_{ij}=\mathbf{\Phi}_i-\mathbf{\Phi}_j.
\end{equation}
We denote by $\mbox{\boldmath$\phi$}_i$ and
$\Delta\mbox{\boldmath$\phi$}_{ij}
\triangleq\mbox{\boldmath$\phi$}_i-\mbox{\boldmath$\phi$}_j$ the
$p-$dimensional vectors obtained by writing $\mathbf{\Phi}_i$ and
$\Delta\mathbf{\Phi}_{ij}$ in lexicographic ordering and
$\mathcal{P}=\{1,\ldots,p\}$ the set of indices corresponding to the
image support. We introduce the following definition.\\
\textbf{Definition} Images $\mathbf{J}_i$ and $\mathbf{J}_j$ are pixel-wise dissimilar
if $\forall k\in \mathcal{P}$, $\Delta\mbox{\boldmath$\phi$}_{ij}(k)
\sim U[0, 2\pi)$.\\
Not surprisingly, nature is replete with images exemplifying
Definition 1. This, in turn, makes it possible to set up a naive
image-based random generator. To confirm this, we used more than
$70,000$ pairs of image patches of resolution $200\times 200$ randomly
extracted from natural images \cite{FreKonEin2007}. For each pair,
we computed $\Delta\mbox{\boldmath$\phi$}_{ij}$ and formulated the
following null hypothesis
\begin{itemize}
\item
$H_0$: $\forall k \in
\mathcal{P}\,\,\Delta\mbox{\boldmath$\phi$}_{ij}(k) \sim U[0,
2\pi)$.
\end{itemize}
which was tested using the Kolmogorov-Smirnov test
\cite{PapPil2004}. For a significance level equal to $0.01$, the
null hypothesis was accepted for $94.05\%$ of the image pairs with
mean $p$-value equal to $0.2848$. In a similar setting, we tested
Matlab's random generator. The null hypothesis was accepted for
$99.48\%$ of the cases with mean $p$-value equal to $0.501$. Fig.
\ref{Fig1} (a)-(b) show a typical pair of image patches considered
in our experiment. Fig. \ref{Fig1} (c) and (d) plot the histograms
of the gradient orientation differences and 40,000 samples drawn
from Matlab's random number generator respectively.

\section{PCA of gradient orientations}\label{S:PCA_GRAD}

\subsection{Cosine-based correlation of gradient orientations}

Given the set of our images $\{\mathbf{I}_i\}$, we compute the
corresponding set of orientation images $\{\mathbf{\Phi}_i\}$ and
measure image correlation using the cosine kernel
\begin{equation}
s(\mbox{\boldmath$\phi$}_i, \mbox{\boldmath$\phi$}_j)
\triangleq\sum_{k\in\mathcal{P}}\cos[\Delta\mbox{\boldmath$\phi$}_{ij}(k)]=cN(\mathcal{P})
 \label{OC}
\end{equation}
where $c\in[-1,1]$. Notice that for highly spatially correlated
images $\Delta\mbox{\boldmath$\phi$}_{ij}(k) \approx 0$ and $c
\rightarrow 1$.

Assume that there exists a subset $\mathcal{P}_2\subset\mathcal{P}$
corresponding to the set of pixels corrupted by outliers. For
$\mathcal{P}_1 = \mathcal{P}-\mathcal{P}_2$, we have
\begin{equation}
s_1(\mbox{\boldmath$\phi$}_i, \mbox{\boldmath$\phi$}_j)
=\sum_{k\in\mathcal{P}_1}\cos[\Delta\mbox{\boldmath$\phi$}_{ij}(k)]=c_1N(\mathcal{P}_1)\label{OC1}
\end{equation}
where $c_1\in[-1,1]$.

\begin{figure}[ht!]
\centering
\includegraphics[width=0.75\columnwidth]{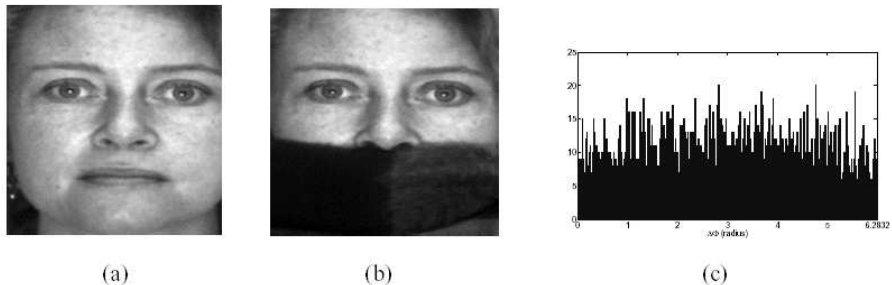}
\caption{(a)-(b) An image pair used in our experiments. (c) The
distribution  of $\Delta\mbox{\boldmath$\phi$}$ for the part of face
occluded by the scarf.}\label{Fig2}
\end{figure}

Not unreasonably, we assume that in $\mathcal{P}_2$, the images are
pixel-wise dissimilar according to Definition 1. For example, Fig.
\ref{Fig2} (a)-(b) show an image pair where $\mathcal{P}_2$ is the
part of the face occluded by the scarf and Fig. \ref{Fig2} (c) plots
the distribution of $\Delta\mbox{\boldmath$\phi$}$ in
$\mathcal{P}_2$.
Before proceeding for $\mathcal{P}_2$, we need the following theorem
\cite{PapPil2004}.\\
\textbf{Theorem II}\\
Let $u(.)$ be a random process and $u(t) \sim U[0, 2\pi)$ then:
 \begin{itemize}
 \item  $\mathbb{E}[\int_{\mathcal{X}} \cos u(t) dt]=0$  for any non-empty interval $\mathcal{X}$  of
$\Re$.
 \item  If $u(.)$ is mean ergodic, then  $\int_{\mathcal{X}} \cos u(t) dt = 0$.
\end{itemize}
We also make use of the following approximation
\begin{equation}
\int_{\mathcal{X}}\cos [\Delta\mbox{\boldmath$\phi$}_{ij}(t)]dt
\approx \sum_{k\in \mathcal{P}} \cos
[\Delta\mbox{\boldmath$\phi$}_{ij}(k)] \label{approx1}
\end{equation}
where with some abuse of notation,
$\Delta\mbox{\boldmath$\phi$}_{ij}$ is defined in the continuous
domain on the left hand side of (\ref{approx1}). Completely
analogously, the above theorem and approximation hold for the case
of the sine kernel.

Using the above results, for $\mathcal{P}_2$, we have
\begin{equation}
s_2(\mbox{\boldmath$\phi$}_i, \mbox{\boldmath$\phi$}_j) = \sum_{k\in
\mathcal{P}_2}\cos[\Delta \mbox{\boldmath$\phi$}_{ij}(k)] \simeq
0\label{OCP2}
\end{equation}
It is not difficult to verify that $\ell_2$-based correlation i.e.
the inner product between two images will be zero if and only if the
images have interchangeably black and white pixels. Our analysis and
(\ref{OCP2}) show that cosine-based correlation of gradient
orientations allows for a much broader class of uncorrelated images.
Overall, unlike $\ell_2$-based correlation where the contribution of
outliers can be arbitrarily large, $s(.)$ measures correlation as
$s(\mbox{\boldmath$\phi$}_i, \mbox{\boldmath$\phi$}_j) =
s_1(\mbox{\boldmath$\phi$}_i,
\mbox{\boldmath$\phi$}_j)+s_2(\mbox{\boldmath$\phi$}_i,
\mbox{\boldmath$\phi$}_j) \simeq c_1N(\mathcal{P}_1)$, i.e. the
effect of outliers is approximately canceled out.

\subsection{The principal components of image gradient orientations}

To show how (\ref{OC}) can be used as a basis for PCA, we first
define the distance
\begin{equation}
d^2(\mbox{\boldmath$\phi$}_i,\mbox{\boldmath$\phi$}_j) =
\sum_{k=1}^p\{1-\cos[\Delta\mbox{\boldmath$\phi$}_{ij}(k)]\}\label{dOC}
\end{equation}
We can write (\ref{dOC}) as follows
\begin{eqnarray}\label{E:L2_COSINE}
d^2(\mbox{\boldmath$\phi$}_i,\mbox{\boldmath$\phi$}_j)& = &\frac{1}{2}\sum_{k=1}^p\{2-2\cos[\mbox{\boldmath$\phi$}_i(k)-\mbox{\boldmath$\phi$}_j(k)]\}\nonumber\\
&=&\frac{1}{2}|| e^{j\mbox{\boldmath$\phi$}_i} -
e^{j\mbox{\boldmath$\phi$}_j} ||^2
\end{eqnarray}
where $e^{j\mbox{\boldmath$\phi$}_i} =
[e^{\mbox{\boldmath$\phi$}_i(1)}, \ldots
,e^{\mbox{\boldmath$\phi$}_i(p)}]^T$. The last equality makes the
basic computational module of our scheme apparent. We define the
mapping from $[0,2\pi)^p$ onto a subset of complex sphere with
radius $\sqrt{N(\mathcal{P})}$
\begin{equation}\label{E:FIRST_MAP}
\mathbf{z}_i(\mbox{\boldmath$\phi$}_i) =
e^{j\mbox{\boldmath$\phi$}_i}
\end{equation}
and apply linear complex PCA to the transformed data $\mathbf{z}_i$.

Using the results of the previous subsection, we can remark the
following\\
\textbf{Remark I} If $\mathcal{P} = \mathcal{P}_1 \cup \mathcal{P}_2$
with $\Delta\mbox{\boldmath$\phi$}_{ij}(k) \sim U[0, 2\pi),\,\,
\forall k\in\mathcal{P}_2$, then
$\mbox{Re}[\mathbf{z}_i^{H}\mathbf{z}_j] \simeq c_1N(\mathcal{P}_1)$\\
\textbf{Remark II}
If $\mathcal{P}_2=\mathcal{P}$, then
$\mbox{Re}[\mathbf{z}_i^{H}\mathbf{z}_j] \simeq 0$ and
$\mbox{Im}[\mathbf{z}_i^{H}\mathbf{z}_j] \simeq 0$.

Further geometric intuition about the mapping $\mathbf{z}_i$ is
provided by the chord between vectors $\mathbf{z}_i$ and
$\mathbf{z}_j$
\begin{equation}\label{E:CHORD}
\mbox{crd}(\mathbf{z}_i,\mathbf{z}_j) = \sqrt{(\mathbf{z}_i
-\mathbf{z}_j)^{H} (\mathbf{z}_i - \mathbf{z}_j)} = \sqrt{2
d^2(\mbox{\boldmath$\phi$}_i,\mbox{\boldmath$\phi$}_j)}
  \end{equation}
Using $\mbox{crd}(.)$, the results of Remark 1 and 2 can be
reformulated as $\mbox{crd}(\mathbf{z}_i,\mathbf{z}_j) \simeq
\sqrt{2((1-c_1) N(\mathcal{P}_1) + N(\mathcal{P}_2))}$ and
$\mbox{crd}(\mathbf{z}_i,\mathbf{z}_j) \simeq
\sqrt{2N(\mathcal{P})}$ respectively.

Overall, $\textbf{Algorithm 1}$ summarizes the steps of our PCA of
gradient orientations.
\newline
\newline
\textbf{Algorithm 1. }\textit{Estimating the principal
subspace}\newline\textbf{Inputs:} A set of $n$ orientation images
$\mathbf{\Phi}_i\textrm{, }i=1,\dots,n$ of $p$ pixels and the number
$k$ of principal components.
\newline\textbf{Step 1.} Obtain $\mbox{\boldmath$\phi$}_i$ by writing $\mathbf{\Phi}_i$ in lexicographic ordering.
\newline\textbf{Step 2.} Compute  $\mathbf{z}_i=e^{j\mbox{\boldmath$\phi$}_i}$, form the matrix of the transformed data
$\mathbf{Z}=[\mathbf{z}_1|\cdots| \mathbf{z}_n]\in
\mathcal{C}^{p\times n}$ and compute the matrix
$\mathbf{T}=\mathbf{Z}^H\mathbf{Z}\in \mathcal{R}^{n\times n}$.
\newline\textbf{Step 3.} Compute the eigen-decomposition of $\mathbf{T}=\mathbf{U}\mathbf{\Lambda}\mathbf{U}^H$
and denote by $\mathbf{U}_k\in \mathcal{C}^{p\times k}$ and
$\mathbf{\Lambda}_k\in \mathcal{R}^{k\times k}$ the $k-$reduced set.
Compute the principal subspace from
$\mathbf{B}_k=\mathbf{Z}\mathbf{U}_k\mathbf{\Lambda}_k^{-\frac{1}{2}}\in
\mathcal{C}^{p\times k}$.
\newline\textbf{Step 4.} Reconstruct using
$\mathbf{\tilde{Z}}=\mathbf{B}_k\mathbf{B}_k^H\mathbf{Z}$.
\newline\textbf{Step 5.} Go back to the orientation domain using $\mathbf{\tilde{\Phi}}=\angle{\mathbf{\tilde{Z}}}$.

Let us denote by $\mathcal{Q} = \{1,\ldots,n\}$ the set of image
indices and $\mathcal{Q}_i$ any subset of $\mathcal{Q}$. We can
conclude the following\\
\textbf{Remark III} If $\mathcal{Q} = \mathcal{Q}_1 \cup
\mathcal{Q}_2$ with $\mathbf{z}_i^{H}\mathbf{z}_j \simeq 0$ $\forall
i\in \mathcal{Q}_2$, $ \forall j \in \mathcal{Q}$ and $i\neq j$,
then, $\exists$ eigenvector $\mathbf{b}_l$ of $\mathbf{B}_n$ such
that $ \mathbf{b}_l \simeq  \frac{1}{N(\mathcal{P})} \mathbf{z}_i $.

A special case of Remark III is the following\\
\textbf{Remark IV} If $\mathcal{Q} = \mathcal{Q}_2$, then
$\frac{1}{N(\mathcal{P})} \mathbf{\Lambda} \simeq
\mathbf{I}_{n\times n}$ and  $\mathbf{B}_n \simeq
\frac{1}{N(\mathcal{P})} \mathbf{Z}$.

\begin{figure}[h]
\centering
\includegraphics[width=0.5\columnwidth]{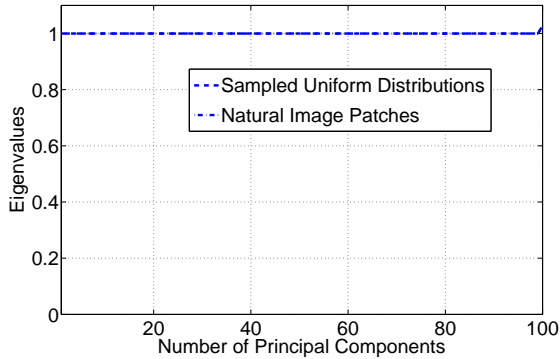}
\caption{The eigen-spectrum of natural images and the eigen-spectrum
of samples drawn from Matlab's random number generator.}\label{Fig3}
\end{figure}

To exemplify Remark IV, we computed the eigen-spectrum of 100 natural
image patches. In a similar setting, we computed the eigen-spectrum
of samples drawn from Matlab's random number generator. Fig.
\ref{Fig3} plots the two eigen-spectrums.

Finally, notice that our framework also enables the direct embedding
of new samples. $\textbf{Algorithm}$ 2 summarizes the
procedure.\newline \textbf{Algorithm 2. }\textit{Embedding of new
samples}\newline\textbf{Inputs:} An orientation image
$\mathbf{\Theta}$ of $p$ pixels and the principal subspace
$\mathbf{B}_k$ of $\textbf{Algorithm 1}$.
\newline\textbf{Step 1.} Obtain $\mbox{\boldmath$\theta$}$ by writing $\mathbf{\Theta}$ in
lexicographic ordering.
\newline\textbf{Step 2.} Compute  $\mathbf{z}=e^{j\mbox{\boldmath$\theta$}}$ and reconstruct using
$\mathbf{\tilde{z}}=\mathbf{B}_k\mathbf{B}_k^H\mathbf{z}$.
\newline\textbf{Step 3.} Go back to the orientation domain using
$\mbox{\boldmath$\tilde{\theta}$}=\angle{\mathbf{\tilde{z}}}$.

\section{Results}\label{S:Results}

\subsection{Face reconstruction}

The estimation of a low-dimensional subspace from a set of a
highly-correlated images is a typical application of PCA
\cite{KirSir1990}. As an example, we considered a set of 50 aligned
face images of image resolution $192\times 168$ taken from the Yale
B face database \cite{GeoBelKri2001}. The images capture the face of
the same subject under different lighting conditions. This setting
usually induces cast shadows as well as other specularities. Face
reconstruction from the principal subspace is a natural candidate
for removing these artifacts.

We initially considered two versions of this experiment. The first
version used the set of original images. In the second version,
20$\%$ of the images was artificially occluded by a $70\times 70$
``Baboon'' patch placed at random spatial locations. For both
experiments, we reconstructed pixel intensities and gradient
orientations with $\ell_2$ PCA and PCA of gradient orientations
respectively using the first 5 principal components.

\begin{figure}[htb!]
\centering
\includegraphics[width=0.75\columnwidth]{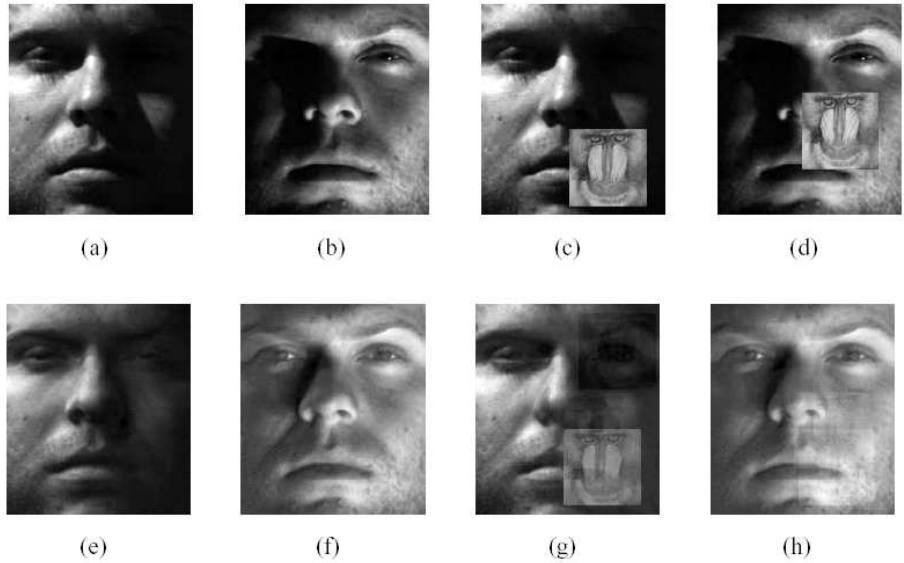}
\caption{PCA-based reconstruction of pixel intensities. (a)-(b)
Original images used in version 1 of our experiment. (c)-(d)
Corrupted images used in version 2 of our experiment. (e)-(f)
Reconstruction of (a)-(b) with 5 principal components. (g)-(h)
Reconstruction of (c)-(d) with 5 principal components. }\label{Fig4}
\end{figure}

\begin{figure}[htb!]
\centering
\includegraphics[width=0.75\columnwidth]{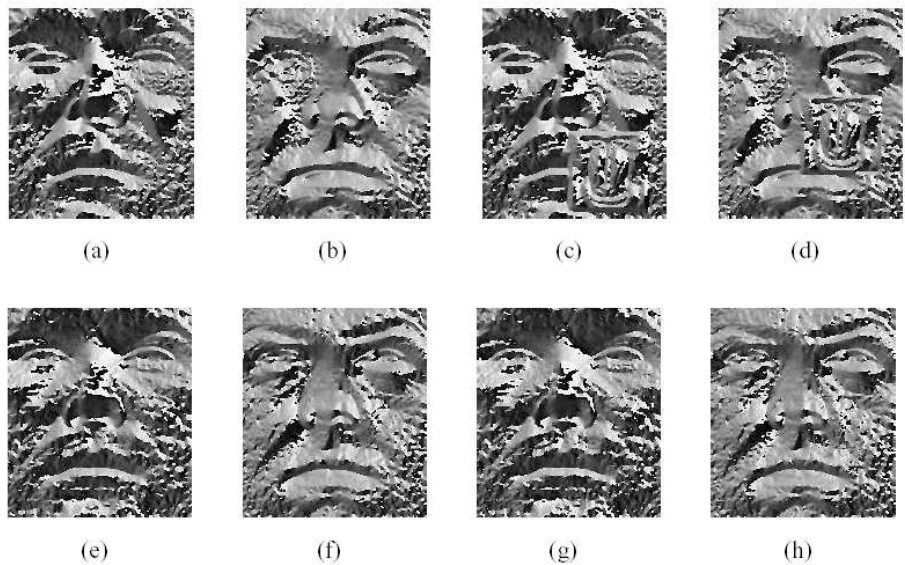}
\caption{PCA-based reconstruction of gradient orientations. (a)-(b)
Original orientations used in version 1 of our experiment. (c)-(d)
Corrupted orientations used in version 2 of our experiment. (e)-(f)
Reconstruction of (a)-(b) with 5 principal components. (g)-(h)
Reconstruction of (c)-(d) with 5 principal components. }\label{Fig5}
\end{figure}

Fig. \ref{Fig4} and Fig. \ref{Fig5} illustrate the quality of
reconstruction for 2 examples of face images considered in our
experiments. While PCA-based reconstruction of pixel intensities is
visually appealing in the first experiment, Fig. \ref{Fig4} (g)-(h)
clearly illustrate that, in the second experiment, the
reconstruction suffers from artifacts. In contrary, Fig. \ref{Fig5}
(e)-(f) and (g)-(h) show that PCA-based reconstruction of gradient
orientations not only reduces the effect of specularities but also
reconstructs the gradient orientations corresponding to the ``face''
component only.

\begin{figure}[htb!]
\centering
\includegraphics[width=0.75\columnwidth]{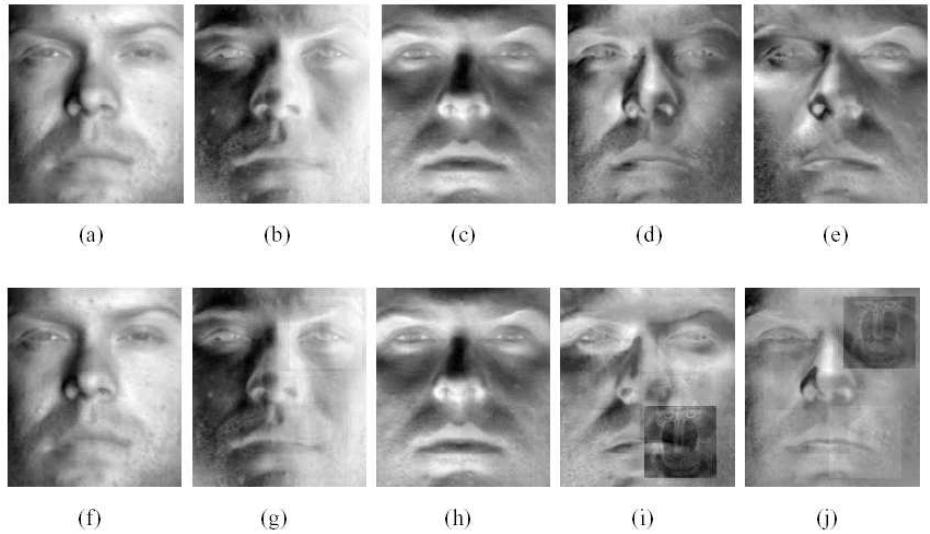}
\caption{The 5 principal components of pixel intensities for (a)-(e)
version 1 and (f)-(j) version 2 of our experiment.}\label{Fig6}
\end{figure}

\begin{figure}[htb!]
\centering
\includegraphics[width=0.75\columnwidth]{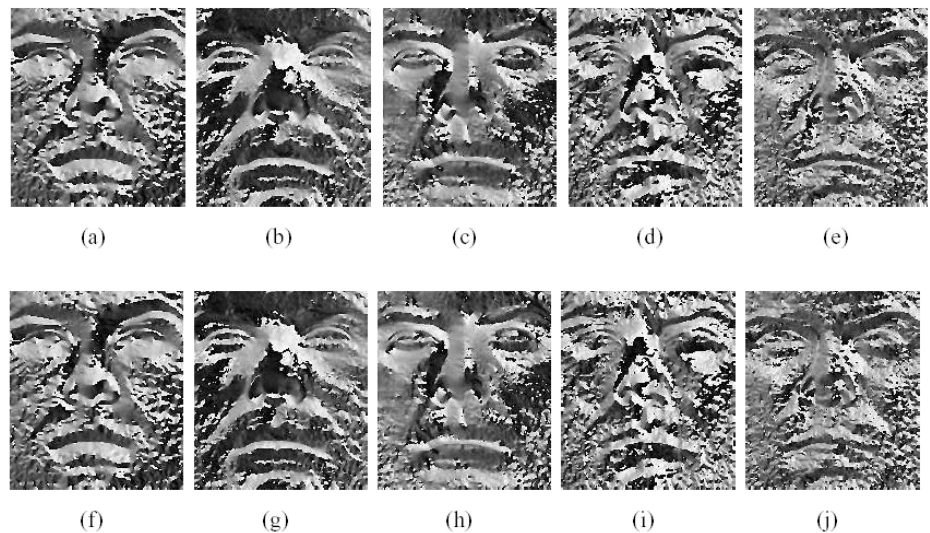}
\caption{The 5 principal components of gradient orientations for
(a)-(e) version 1 and (f)-(j) version 2 of our
experiment.}\label{Fig7}
\end{figure}

This performance improvement becomes more evident by plotting the
principal components for each method and experiment. Fig. \ref{Fig6}
shows the 5 dominant Eigenfaces of $\ell_2$ PCA. Observe that, in
the second experiment, the last two Eigenfaces (Fig. \ref{Fig6} (i)
and (j)) contain ``Baboon'' ghosts which largely affect the quality
of reconstruction. In contrary, a simple visual inspection of Fig.
\ref{Fig7} reveals that, in the second experiment, the principal
subspace of gradient orientations (Fig. \ref{Fig7} (f)-(j)) is
artifact-free which in turn makes dis-occlusion in the orientation
domain feasible.

\begin{figure}[htb!]
\centering
\includegraphics[width=0.75\columnwidth]{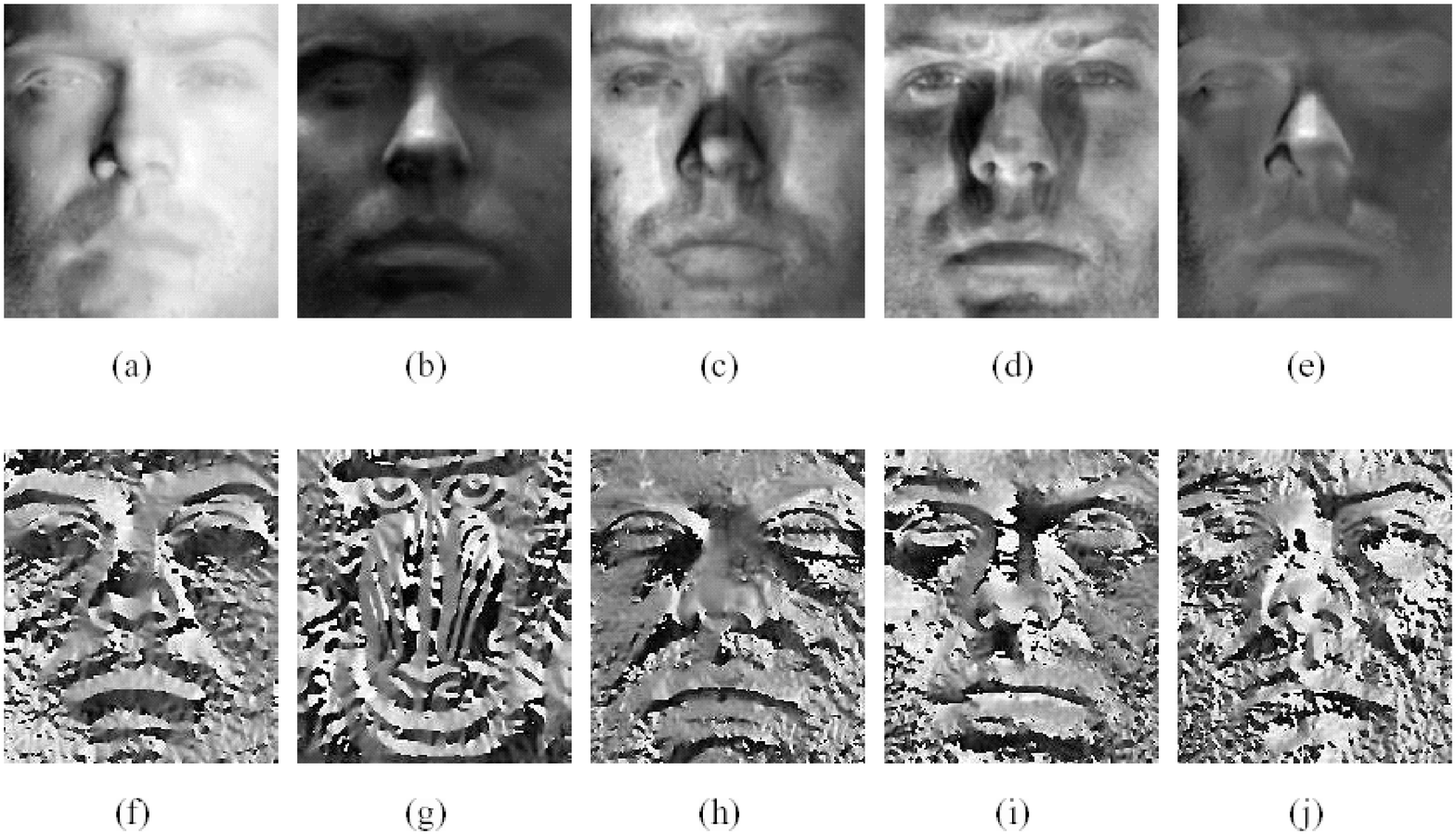}
\caption{(a)-(e) The 5 principal components of pixel intensities for
version 3 of our experiment and (f)-(j) The 5 principal components
of gradient orientations for the same experiment.}\label{Fig8}
\end{figure}

Finally, to exemplify Remark 3, we considered a third version of our
experiment where 20$\%$ of the images were replaced by \textit{the
same} $192\times 168$ ``Baboon'' image. Fig. \ref{Fig8} (a)-(e) and
(f)-(j) illustrate the principal subspace of pixel intensities and
gradient orientations respectively. Clearly, we may observe that
$\ell_2$ PCA was unable to handle the extra-class outlier. In
contrary, PCA of gradient orientations successfully separated the
``face'' from the ``Baboon'' subspace i.e. no eigenvectors were
corrupted by the ``Baboon'' image. Note that the ``face'' principal
subspace is not the same as the one obtained in versions 1 and 2.
This is because only 80$\%$ of the images in our dataset was used in
this experiment.

\section{Conclusions}\label{S:Conclusions}

We introduced a new concept: PCA of gradient orientations. Our
framework is as simple as standard $\ell_2$ PCA, yet much more
powerful for efficient subspace-based data representation. Central
to our analysis is the distribution of gradient orientation
differences and the cosine kernel which provide us a consistent way
to measure image dissimilarity. We showed how this dissimilarity
measure can be naturally used to formulate a robust version of PCA.
Extensions of our scheme
span a wide range of theoretical topics and applications; from
statistical machine learning and clustering to object recognition
and tracking.

%


\bibliographystyle{IEEEbib}

\end{document}